\documentclass{article}

\usepackage{microtype}
\usepackage{graphicx}
\usepackage{subfigure}
\usepackage{booktabs} % for professional tables

\usepackage{hyperref}

\usepackage[accepted]{icml2020}

\usepackage{bm}
\usepackage{mathtools}
\usepackage{amsmath}
\usepackage{amsfonts}

\usepackage{tikz,pgfplots}

\usepackage{xcolor}

\icmltitlerunning{Causal Discovery by Kernel Deviance Measures with Heterogeneous Transforms}

\begin{document}

\twocolumn[
\icmltitle{Causal Discovery by Kernel Deviance Measures with Heterogeneous Transforms}



\icmlsetsymbol{equal}{*}

\begin{icmlauthorlist}
\icmlauthor{Tim Tse}{h}
\icmlauthor{Zhitang Chen}{h}
\icmlauthor{Shengyu Zhu}{h}
\icmlauthor{Yue Liu}{h}
% \icmlauthor{Iaesut Saoeu}{ed}
% \icmlauthor{Fiuea Rrrr}{to}
% \icmlauthor{Tateu H.~Yasehe}{ed,to,goo}
% \icmlauthor{Aaoeu Iasoh}{goo}
% \icmlauthor{Buiui Eueu}{ed}
% \icmlauthor{Aeuia Zzzz}{ed}
% \icmlauthor{Bieea C.~Yyyy}{to,goo}
% \icmlauthor{Teoau Xxxx}{ed}
% \icmlauthor{Eee Pppp}{ed}
\end{icmlauthorlist}

\icmlaffiliation{h}{Huawei Noah's Ark Lab}
% \icmlaffiliation{goo}{Googol ShallowMind, New London, Michigan, USA}
% \icmlaffiliation{ed}{School of Computation, University of Edenborrow, Edenborrow, United Kingdom}

\icmlcorrespondingauthor{Tim Tse}{tim.tse@huawei.com}
\icmlcorrespondingauthor{Zhitang Chen}{chenzhitang2@huawei.com}
\icmlcorrespondingauthor{Shengyu Zhu}{zhushengyu@huawei.com}
\icmlcorrespondingauthor{Yue Liu}{liuyue52@huawei.com}

\icmlkeywords{Machine Learning, ICML}

\vskip 0.3in
]



\printAffiliationsAndNotice{}  % leave blank if no need to mention equal contribution

\begin{abstract}
The discovery of causal relationships in a set of random variables is a fundamental objective of science and has also recently been argued as being an essential component towards real machine intelligence. One class of causal discovery techniques are founded based on the argument that there are inherent structural asymmetries between the causal and anti-causal direction which could be leveraged in determining the direction of causation. To go about capturing these discrepancies between cause and effect remains to be a challenge and many current state-of-the-art algorithms propose to compare the norms of the kernel mean embeddings of the conditional distributions. In this work, we argue that such approaches based on RKHS embeddings are insufficient in capturing principal markers of cause-effect asymmetry involving higher-order structural variabilities of the conditional distributions. We propose Kernel Intrinsic Invariance Measure with Heterogeneous Transform (KIIM-HT) which introduces a novel score measure based on heterogeneous transformation of RKHS embeddings to extract relevant higher-order moments of the conditional densities for causal discovery.
%We build on-top of these class of methods where specifically, we identify the limitations of a current technique which relies on computing an eigendecomposition of the embeddings. We show that a global linear projection is insufficient in capturing information in the conditional distributions which vary across each instance. Motivated by these observations, we propose a novel local nonlinear transformation in the form of an artificial neural network which instance-wise, extracts high-order level statistics that are relevant to causal discovery.
Inference is made via comparing the score of each hypothetical cause-effect direction. Tests and comparisons on a synthetic dataset, a two-dimensional synthetic dataset and the real-world benchmark dataset T\"{u}bingen Cause-Effect Pairs verify our approach. In addition, we conduct a sensitivity analysis to the regularization parameter to faithfully compare previous work to our method and an experiment with trials on varied hyperparameter values to showcase the robustness of our algorithm.
\end{abstract}

\section{Introduction}
The detection of causal relationships is important in science as it allows one to infer the results of an outcome subjected to different interventions \cite{Pearl:2009:CMR:1642718}. For example, it is common in the medical sciences to determine the genetic causation of certain diseases while in the social sciences one might seek to understand the driving factors that determine socioeconomic status \cite{doi:10.1086/662659}. Historically, the majority of causal relationships are discovered through randomized controlled experiments but constraints such as financial and ethical limitations have motivated research into methodologies that operate solely on observational data \cite{Mitrovic:2018:CIV:3327757.3327802}.

In more recent times, causality has also been gaining more traction in the machine learning (ML) community. The recent advent of Deep Learning (DL) \cite{article_deep_learning,Goodfellow-et-al-2016} has had a remarkable impact, with DL-based models achieving state-of-the-art results in tasks ranging from natural language processing and speech recognition to computer vision and reinforcement learning. Despite these successes, there remains significant challenges to be addressed such as robustness to adversarial attacks, the demand for enormous amounts of labelled data and the lack of generalization across domains \cite{Waldrop1074}. The application of causality to redress these limitations appear to be promising as the interactions between random variables could be leveraged as a prior to render an algorithm more robust to adversarial attacks, data-efficient and generalize better to unseen domains where an otherwise ``associative method'' could not.

Some examples of recent work in this direction include \cite{DBLP:journals/corr/abs-1901-08162} which introduces a model-free meta-reinforcement learning algorithm that learns the causal structure of the environment thereby providing the agent with the ability to perform and interpret experiments. \cite{DBLP:journals/corr/abs-1902-03380} proposes a causal inference framework for robust visual reasoning via do-calculus, treating adversarial attacks as interventions. Finally, the work of \cite{article123123} propounds Causal Domain Adaptation Networks, an artificial network architecture which makes use of causal representation of joint distributions for the task of domain adaptation which essentially involves understanding and making use of distribution changes across domains. We refer the reader to \cite{2019arXiv191110500S} for an in-depth discussion as well as addition examples on the topic of causality for machine learning.

How to actually go about learning the causal relationships of r.v.s is itself a broad topic but in general, the various methods could be roughly grouped based on their common theme. A class of methods going by the name of structural equation modelling (SEM) \cite{doi:10.1146/annurev-statistics-031017-100630,Shimizu:2006:LNA:1248547.1248619} seek to model structural relationships through graphical models but are limited by their ability to identify models only up to a Markov equivalence class. Another group of methods \cite{book_sprites,Sun:2007:KCL:1273496.1273604} attempt to infer the causal structure by testing for independence of the conditional distributions of the data but are not robust to the choice of conditional independence test. A final collection of techniques \cite{NIPS2008_3548,Mooij:2009:RDM:1553374.1553470,Zhang:2009:IPC:1795114.1795190} conjecture that there are innate asymmetries between two interacting r.v.s. which could be leveraged for causal discovery but are limited by the assumption of a particular functional form and noise model.

A particular interesting interpretation of cause-effect asymmetry argues that nature, tends from order to disorder in accordance with the second law of thermodynamics, suggesting that one may be able to infer the sequence of causation in a set of events by determining asymmetries in their measured complexity \cite{Janzing_2016}. From a statistical point of view, one interpretation of asymmetry in cause-effect \cite{inproceedings1234235234} presents itself in the idea that the factorization of the joint distribution $p(X, Y)$ is simpler in the causal direction $X \rightarrow Y$ as opposed to the anti-causal direction $Y \rightarrow X$, i.e. \begin{equation}\label{kolmogorov}\mathcal{K}(p(X)) + \mathcal{K}(p(Y|X)) \leq \mathcal{K}(p(Y)) + \mathcal{K}(p(X|Y)) \end{equation} where $\mathcal{K}(\cdot)$ denotes the Kolmogorov complexity (KC) \cite{DBLP:journals/corr/abs-0809-2754}. Despite its theoretical appeal, it is well-known that the KC measure is essentially uncomputable and as a workaround, metrics such as Minimal Description Length \cite{Lemeire2006CausalMA} and RKHS norms \cite{12341234,Sun:2008:CRE:1352927.1352998} have been proposed as tractable substitutes. A recent work of this class \cite{Mitrovic:2018:CIV:3327757.3327802} proposes a score-based measure based on the variance of the RKHS norms while \cite{TODO_Zhitang_arxiv} propounds an extension amounting to a linear projection of the norms of the variance between the RKHS embeddings.

Our work builds on top of state-of-the-art approaches based on RKHS embeddings. In this paper, we argue that metrics of current methods are insufficient in capturing structural variabilities of the conditional distributions which are vital information for inferring cause-effect asymmetry. We instead propose Kernel Instrinsic Invariance Measure with Heterogeneous Transforamtion (KIIM-HT) which presents a novel score measure based on heterogeneous transformation of RKHS embeddings to extract relevant higher-order deviances of the conditional distributions for causal discovery.

The remainder of the paper is organized as follows: first we provide some preliminaries then we discuss some existing work in causal discovery. Next, we describe our proposed method followed by experiments on a synthetic dataset, a two-dimensional synthetic dataset, the  T\"{u}bingen Cause-Effect Pairs (TCEP) dataset, a sensitivity analysis to the regularization parameter and a demonstration of the robustness of our method to different hyperparameters and finally, we will conclude.
\section{Preliminaries}\label{preliminaries}
We briefly review some concepts on kernel methods and kernel mean embeddings, which form the bedrock of our method and few other related works.

Kernel methods are a class of machine learning algorithms that work by mapping an input space $\mathcal{X}$ to a possibly infinite dimensional space $\mathcal{H}$ where certain operations in $\mathcal{X}$ could be expressed as dot products in $\mathcal{H}$. This so-called kernel trick allows one to bypass the explicit and sometimes intractable (i.e., infinite) representation of coordinates in $\mathcal{X}$ which are instead mapped into $\mathcal{H}$, $\bm{x} \rightarrow \phi(\bm{x}) \coloneqq  k(\bm{x}, \cdot)$ via positive definite kernel $k(\cdot, \cdot)$, satisfying $\langle \phi(\bm{x}), \phi(\bm{y}) \rangle_\mathcal{H} = k(\bm{x}, \bm{y})$. Effectively, (non)linear transformations and dot product operations are replaced by kernel evaluations which, in one possible interpretation, is a pairwise similarity measure of the inputs.

Given a probability distribution $p(X)$ on $\mathcal{X}$, the kernel mean embedding \cite{6530747}  of $p(X)$ in $\mathcal{H}_{\mathcal{X}}$ is given by $$\mu_X \coloneqq \mathbb{E}_X[k(X, \cdot)] = \mathbb{E}_X[\phi(X)] = \int_\mathcal{X} \phi(\bm{x})p(\bm{x})d \bm{x}.$$ Oftentimes, we do not have access to the underlying true distribution and thus calls for an estimation of the embedding with finte samples. Given samples $\{\bm{x}_i|i=1,\dots,n\}$ the empirical estimate is given by $$\hat{\mu}_X = \frac{1}{n}\sum_{i=1}^n k(\bm{x}_i, \cdot).$$ \cite{Scholkopf:2001:LKS:559923} show that if $k$ is a characteristic kernel then the mapping is injective. An example of such a kernel is the radial basis function (RBF) kernel defined as $$k(\bm{x}, \bm{y}) = \sigma_f^2 \exp\bigg(-\frac{\left\lVert \bm{x} - \bm{y}\right\rVert_2^2}{2 \ell^2}\bigg),$$ where $\ell$ is the length-scale and $\sigma_f$ is the signal amplitude. Similarly, given a conditional distribution $p(X|Y = \bm{y})$ instantiated at $\bm{y}$, the kernel conditional mean embedding is defined as $$\mu_{X|\bm{y}} \coloneqq \mathbb{E}_{X|\bm{y}}[\phi(X)] = \int_{\mathcal{X}} \phi(\bm{x}) p(\bm{x} | \bm{y}) d \bm{x}$$ and likewise, the corresponding empirical estimate is given by $$\hat{\mu}_{X|\bm{y}} = \bm{\Psi}(\bm{K}_y + \lambda \bm{I})^{-1} \bm{k}_{\bm{y}}$$ where $\bm{K}_y$ is the gram matrix (i.e., $[\bm{K}_y]_{i,j} = k(\bm{y}_i, \bm{y}_j)$), $\bm{\Psi} = [\psi(\bm{x}_1),\dots,\psi(\bm{x}_n)]$ and $\bm{k}_{\bm{y}} = [k(\bm{y}, \bm{y}_1),\dots,k(\bm{y},\bm{y}_n)]^\top$.

%Causal discovery routinely calls for the comparison of conditional distributions which their kernel embedding is defined as $$\mu_{Y|\bm{x}} \coloneqq \mathbb{E}_{Y|\bm{x}}[\phi(Y)] = \int_{\mathcal{Y}} \phi(\bm{y}) p(\bm{y} | \bm{x}) d \bm{y}$$ and likewise, their empirical estimate is given by $$\hat{\mu}_{Y|\bm{x}} = \bm{\Psi}(\bm{K}_x + \lambda \bm{I})^{-1} \bm{k}_{\bm{x}}$$ where $\bm{K}_x$ is the gram matrix (i.e., $[\bm{K}_x]_{i,j} = k(\bm{x}_i, \bm{x}_j)$), $\bm{\Psi} = [\psi(\bm{x}_1),\dots,\psi(\bm{x}_n)]$ and $\bm{k}_{\bm{x}} = [k(\bm{x}, \bm{x}_1),\dots,k(\bm{x},\bm{x}_n)]^\top$.
\section{Related Work}\label{related_work}
The two prior works that are most relevant to our current research are 1) Kernel Conditional Deviance for Causal Inference (KCDC) and 2) Kernel Intrinsic Invariance Measure (KIIM).

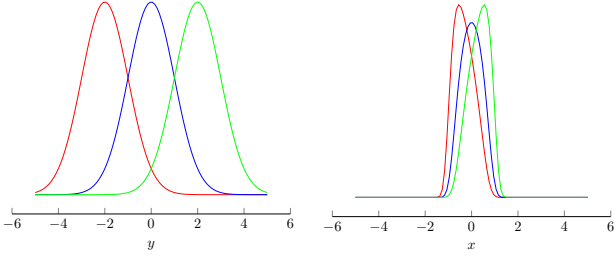
\begin{figure}
\centering
\begin{tikzpicture}[scale=0.54]
\begin{axis}[axis x line*=bottom,hide y axis,xlabel={$y$}]
\addplot[color=red,samples=100]{exp(((x +2)^2 + 0)/-2)/(2 * pi)};
\addplot[color=blue,samples=100]{exp(((x)^2)/-2)/(2 * pi)};
\addplot[color=green,samples=100]{exp(((x - 2)^2 + 0)/-2)/(2 * pi)};
%\addplot[color=blue]{x^2};
\end{axis}
\end{tikzpicture}
%Here ends the first plot
\hskip 5pt
%Here begins the 3d plot
\begin{tikzpicture}[scale=0.54]
\begin{axis}[axis x line*=bottom,hide y axis,xlabel={$x$}]
\addplot[color=red,samples=100]{exp(((-1 - x^3 -x)^2 + x^2)/-2)/(1.5 * pi)};
\addplot[color=blue,samples=100]{exp(((0 - x^3 -x)^2 + x^2)/-2)/(2 * pi)};
\addplot[color=green,samples=100]{exp(((1 - x^3 -x)^2 + x^2)/-2)/(1.5 * pi)};
%\addplot[color=blue]{x^2};
\end{axis}
\end{tikzpicture}
\caption{Conditional distributions of the causal direction, $p(y|x)$, and the anti-causal direction, $p(x|y)$.}
\label{direction}
\end{figure}

As aforementioned, KCDC belongs to the class of causal discovery algorithms that posit an inherent asymmetry between the causal and anti-causal direction. Consider the illustrative toy example, $y = x^3 + x + \epsilon$ where $\epsilon \sim \mathcal{N}(0, 1)$. Figure \ref{direction} depicts the asymmetry between the conditional distributions in the causal direction $p(y|x)$ and the anti-causal direction $p(x|y)$ for different values of $x$ and $y$, respectively. 
 Note the structural variability of the conditional distributions is larger in the anti-causal direction than the causal direction. It is stressed that structural variability does not refer to the variability in location and scale but rather to aspects such as the number of modes and in higher order moments. If one were to imagine the conditional distributions as functions mapping $x$ to $y$ and vice-versa, then the mapping in the causal direction would produce outputs with the same structure for different inputs whereas in the anti-causal direction, the outputs exhibit diverse structural variations across different inputs.

From this insight, KCDC provides a new interpretation of asymmetry based on the KC of the data-generating mechanism. In contrast to other work which measures asymmetry in terms of the KC of the joint distributions (i.e., Equation \ref{kolmogorov}), KCDC proposes instead to interpret this asymmetry via the Kolgomorov complexity of the conditional distribution. More specifically, KCDC argues that their proposed asymmetry metric is realized by the KC of the input distribution being independent of the causal mechanism whereas on the other hand, in the anti-causal direction, there is dependence in the input and the minimum description length of the data-generating mechanism. This (in)dependence could be measured by looking at the variability in KC of the causal mechanism for different inputs. To circumvent the uncomputability of the KC, KCDC resorts to comparing the variability of conditional distributions embedded in RKHS given by $$\mathcal{S}_{X \rightarrow Y} = \frac{1}{n} \sum_{i=1}^n \Big (\big \|\mu_{Y|\bm{x}_i}\big \|_{\mathcal{H}_\mathcal{Y}}-\frac{1}{n}\sum_{j=1}^n\big \|\mu_{X|\bm{y}_j}\big \|_{\mathcal{H}_\mathcal{X}}\Big)^2$$ where $\mathcal{H}_\mathcal{X}$ and $\mathcal{H}_\mathcal{Y}$ are RKHSs defined on $\mathcal{X}$ and $\mathcal{Y}$, respectively and $\mu_{Y|\bm{x}_i}$ is the mean embedding of the conditional distribution $p(\bm{y}|\bm{x})$ evaluated at $\bm{x}_i$. The direction of causation is made via a simple decision rule that compares $\mathcal{S}_{X \rightarrow Y}$ and $\mathcal{S}_{Y \rightarrow X}$.

\cite{TODO_Zhitang_arxiv} brings to attention potential issues with identifying distributions by simply comparing the norms of the embeddings. Specifically, under the mild assumption that domain $\mathcal{X}$ is symmetric about the origin and the entailing kernel is stationary, then given arbitrary densities $p(\bm{x})$ and $q(\bm{x})$ where $p(\bm{x}) = q(-\bm{x})$, it can be shown by applying Bochner's theorem \cite{rudin1990fourier} that $\|\mu_p\|_{\mathcal{H}_{\mathcal{X}}} = \|\mu_q\|_{\mathcal{H}_{\mathcal{X}}}$. This result implies that $p(\bm{x})$ and $q(\bm{x})$ may share the same norm despite exhibiting large structural variability between the two distributions (i.e., for skewed distributions, the structure of $p(\bm{x})$ and $q(\bm{x})$ are different). In general, given an arbitrary probability density, there exists with high probability a different probability density with the same norm in RKHS embedding. Motivated by this observation, they instead propose KIIM which introduces the score in the direction of $X \rightarrow Y$ given by
$$\mathcal{S}_{X \rightarrow Y} = \min_{\bm{W}} \sum_{i=1}^n \big \|\bm{W}^\top \mu_{Y|\bm{x}_i} - \frac{1}{n} \sum_{j=1}^n \bm{W}^\top \mu_{Y|\bm{x}_j} \big \|^2_{\mathcal{H}_\mathcal{Y}}$$ which could be interpreted as the calculation of the difference of the conditional distributions instantiated at different values. The score is zero if and only if all the conditional distributions are equal as a result of the injective property of the kernel mappings. $\bm{W}$ is a projection matrix which maps each conditional distribution into a subspace where the location and scale are removed while leaving the higher-order deviances which are intended for use in quantifying the structural variability of said conditional distributions.
%\indent \cite{TODO_Zhitang_arxiv} points out potential problems in identifying distributions simply by comparing the norms of the embeddings. Namely, they demonstrate with high probability that given an arbitrary probability density, there exists a different probability density with the same norm in RKHS embedding and exemplify a special case of this when two probability densities are a reflection of one another and share a domain that is symmetric w.r.t the origin. Motivated by this observation, they propose Kernel Intrinsic Invariance Measure (KIIM) given by $$\mathcal{S}_{X \rightarrow Y} = \min_{\bm{W}} \sum_{i=1}^n \big \|\bm{W}^\top \mu_{Y|\bm{x}_i} - \frac{1}{n} \sum_{j=1}^n \bm{W}^\top \mu_{Y|\bm{x}_j} \big \|^2_{\mathcal{H}_\mathcal{Y}}$$ which amounts to an eigendecompostition of a kernel matrix task measuring intrinsic deviance/invariance.
%Talk about KCDC in more detail here
%Talk about KIIM in more detail here
\section{Kernel Intrinsic Invariance Measure with Heterogeneous Transformation}\label{method}
%The main impetus behind the methodology of KIIM centers around the argument that a linear transformation on the kernel mean embeddings projects the kernel embeddings to a space spanned by a set of statistics of the conditional distributions. The score is then computed based on the relevant higher-order statistics such as skewness and kurtosis while discarding the confounding lower-order statistics such as mean and variance, thereby allowing the score to capture ``intrinsic deviances/invariances'' in the conditional distributions.\\
Despite the positive results reported, we demonstrate that a global linear projection may not be a sufficient transformation to capture the desired statistics in the conditional distributions. To illustrate this, consider the following simple example. Let $y = x + \epsilon$ with $\epsilon \sim \mathcal{N}(0, 1)$. The conditional distribution evaluated at $x_i$ is given by $$p(y|x_i) = \frac{1}{\sqrt{2 \pi}} \exp\bigg (-\frac{(y-x_i)^2}{2} \bigg).$$ Now, define $p_{Y_i}(y) \coloneqq p(y|x_i)$ and consider the transformation of random variable $Y_i$ given by $Z_j \coloneqq Y_i + b_j$. The pdf of $Z_j$ is given by $$p_{Z_j}(z) = \frac{1}{\sqrt{2 \pi}} \exp\bigg (-\frac{(z-b_j-x_i)^2}{2} \bigg)$$ which implies that an arbitrary conditional distribution $p(y|x_j)$ can be obtain from another conditional distribution $p(y|x_i)$ via the offset $b_j = x_j - x_i$. Now, consider the quadratic kernel $x \mapsto \phi(x) = [1,x,x^2]^\top$, then \begin{equation*}
\begin{split}
\mu_{Z_j} & = \int_{\mathcal{Z}} \phi(z) p_{Z_j}(z) dz \\
 %& = \int_{\mathcal{Y}} \phi(y + b_j) p_{Y_i}(y) dy \\
 %& = \int_{\mathcal{Y}} M(b_j) \phi(y) p_{Y_i}(y) dy \\
 & = M(b_j) \mu_{Y_i}
\end{split}
\end{equation*} where $$ M(b_j) = \begin{bmatrix} 
1 & 0 & 0 \\
b_j & 1 & 0 \\
b_j^2 & 2b_j & 1
\end{bmatrix}.$$ This example demonstrates that even in the simple case where the conditional distributions are offsets of one another and the embedding kernel is a three variable quadratic, $\mu_{Z_j}$ and $\mu_{Y_i}$ are related via a transformation $M(b_j)$ which is a function of the input data-point $b_j$ and therefore, there does not exist a global transformation matrix $\bm{W}$ that captures the statistical relations of all pairwise kernel embedding of the conditional distributions.

Motivated by the previous observations, we propound to rectify the limitations of a global linear projection with a novel score metric based on the heterogeneous projections of the embeddings. Our new algorithm which we call Kernel Intrinsic Invariance Measure with Heterogeneous Transformation (KIIM-HT) introduces the score in the direction of $\bm{x} \rightarrow \bm{y}$ given by \begin{multline}\label{nl_kiim_score}\mathcal{S}_{\bm{x} \rightarrow \bm{y}}^* = \min_{\theta} \frac{2}{n(n-1)}\sum_{i=1}^n \sum_{j=1}^{i-1} \big \|\bm{W}_{\theta}^\top (\bm{x}_i) \mu_{Y|\bm{x}_i} \\ - \bm{W}_{\theta}^\top (\bm{x}_j) \mu_{Y|\bm{x}_j} \big \|_{\mathcal{H}_{\mathcal{Y}}}^2,\end{multline} where $\bm{W}_{\theta}(\cdot)$ is a heterogeneous projection matrix. We leverage the power of artificial neural networks and parameterize $\bm{W}_{\theta}(\cdot)$ as a feedforward neural network with parameters $\theta$. Equation \ref{nl_kiim_score} could be perceived as a pairwise comparison of the kernel mean embeddings of conditional distributions in a subspace and optimizing for $\theta$ amounts to learning a heterogeneous projection into said subspace which preserves the relevant statistics for causal discovery. We underscore the notion that the projection matrix for the $i^{\text{th}}$ instance is a function of the $i^{\text{th}}$ input as motivated by our aforementioned observation of the insufficiency of a global projection.

The empirical estimate of Equation \ref{nl_kiim_score} is given by \begin{equation}\label{nl_kiim_score_empirical}
\hat{\mathcal{S}}_{\bm{x} \rightarrow \bm{y}} = \min_{\theta} \mathcal{S}_{\bm{x} \rightarrow \bm{y}}
\end{equation} where \begin{multline*}
\mathcal{S}_{\bm{x} \rightarrow \bm{y}} = \sum_{i=1}^{n} \sum_{j=1}^{i-1} \big \| \widetilde{\bm{W}}_{\theta}^\top(\bm{x}_i) \bm{K}_y (\bm{K}_x + \lambda \bm{I})^{-1} \bm{k}_{\bm{x}_i} \\- \widetilde{\bm{W}}_{\theta}^\top(\bm{x}_j) \bm{K}_y (\bm{K}_x + \lambda \bm{I})^{-1} \bm{k}_{\bm{x}_j} \big \|_{\mathcal{H}_{\mathcal{Y}}}^2.
\end{multline*} We assume that $\bm{W}_{\theta}(\cdot)$ lies in the span of $\bm{\Psi}$ hence we can write $\widetilde{\bm{W}}_{\theta}(\cdot) = \bm{\Psi}\bm{W}_{\theta}(\cdot)$ and $\bm{K}_y \eqqcolon \bm{\Psi}^\top \bm{\Psi}$ is the gram matrix for the effect (i.e., $[\bm{K}_y]_{i,j} = k(\bm{y}_i, \bm{y}_j)$) while all other parameters remain the same as described in the previous section. We note that a trivial solution exists where $\widetilde{\bm{W}}_{\theta}^\top(\bm{z}) = \bm{0}$ for all $\bm{z}$ and to prevent the optimization routine from converging to this degenerate solution, we introduce a regularization term to Equation \ref{nl_kiim_score_empirical}, resulting in \begin{equation}\label{nl_kiim_score_empirical_reg}
\mathcal{S}_{\bm{x} \rightarrow \bm{y}}^{\text{reg}} = \min_{\theta} \bigg \{ \mathcal{S}_{\bm{x} \rightarrow \bm{y}} + \frac{\lambda_{\text{reg}}}{n} \sum_{i=1}^n \frac{1}{\| \bm{W}_{\theta}(\bm{x}_i)\|_2^2} \bigg \}\end{equation} where $\lambda_{\text{reg}}$ is a hyperparameter which determines the degree of penalization for the trivial solution. We take a straightforward approach in optimizing Loss \ref{nl_kiim_score_empirical_reg} where the score is taken to be the lowest value attained from running a stochastic gradient descent algorithm for $\mathcal{I}$ iterations. Having obtained both $\mathcal{S}_{\bm{x} \rightarrow \bm{y}}$ and $\mathcal{S}_{\bm{y} \rightarrow \bm{x}}$ the direction of causation $D(X, Y)$ is decided via the decision rule \begin{equation}\label{decision_ruel} D(X, Y) = \begin{cases}
      X \rightarrow Y & \text{if } \mathcal{S}_{\bm{x} \rightarrow \bm{y}} < \mathcal{S}_{\bm{y} \rightarrow \bm{x}} \\
      Y \rightarrow X & \text{if } \mathcal{S}_{\bm{x} \rightarrow \bm{y}} > \mathcal{S}_{\bm{y} \rightarrow \bm{x}} \\
      \text{undecided} & \text{if } \mathcal{S}_{\bm{x} \rightarrow \bm{y}} = \mathcal{S}_{\bm{y} \rightarrow \bm{x}}
   \end{cases}.
\end{equation} Algorithm \ref{ouralgo} provides a pseudocode summary of the overall algorithm. \begin{algorithm}
\caption{KIIM-HT Algorithm}
\label{ouralgo}
\hspace*{1.4em} \textbf{Input:} Realizations $\{(\bm{x}_i, \bm{y}_i)\}_{i=1}^n$ of $(X ,Y)$, regularization hyperparamter $\lambda$, number of SGD iterations $\mathcal{I}$\\
\hspace*{1.1em} \textbf{Output:} Causal direction $(X \rightarrow Y$ or $Y \rightarrow X)$
\begin{algorithmic}[1]
    \STATE Compute gram matrices $\bm{K}_x$ and $\bm{K}_y$
    \STATE Compute $\bm{K}_y (\bm{K}_x + \lambda \bm{I})^{-1} \bm{k}_{\bm{x}_i}$ for $i = 1,\dots,n$
    \STATE Minimize Loss \ref{nl_kiim_score_empirical_reg} via $\mathcal{I}$ iterations of SGD
    \STATE Take $\mathcal{S}_{\bm{x} \rightarrow \bm{y}}$ to be the lowest value encountered in the $\mathcal{I}$ iterations of SGD
    \STATE Repeat steps 1 through 4 for the other direction
    \STATE Determine causal direction $D(X, Y)$ using Equation \ref{decision_ruel}
    \STATE \textbf{return} $D(X, Y)$
\end{algorithmic}
\end{algorithm}

KIIM adopts a importance re-weighting scheme motivated by the observation that real-world data exhibits noise and outliers that ought to be re-weighted to minimize their detrimental effects. The re-weighting scheme is given  as \begin{equation*}
\begin{split}
\mathcal{C}_{YX}^{\text{ref}} & \coloneqq \int_{\mathcal{X}} \phi(\bm{y}) \otimes \phi(\bm{x}) p(\bm{y} | \bm{x}) u(\bm{x}) d \bm{x},\\
\mathcal{C}_{XX}^{\text{ref}} & \coloneqq \int_{\mathcal{X}} \phi(\bm{x}) \otimes \phi(\bm{x})  d \bm{x}
\end{split}
\end{equation*} and $$\mu_{Y | \bm{x}}^{\text{ref}} = \mathcal{C}_{YX}^{\text{ref}} \big (\mathcal{C}_{YX}^{\text{ref}} \big )^{-1} \phi(\bm{x})$$ where $u(\bm{x})$ is the reference distribution. This re-weighting scheme could naturally be extended to KIIM-HT by substituting the empirical estimation with $$\hat{\mu}_{Y|\bm{x}}^{\text{ref}} = \bm{\Psi}\bm{H}\bm{R}^{\frac{1}{2}}  (\bm{R}^{\frac{1}{2}}\bm{H}\bm{K}_x \bm{H}\bm{R}^{\frac{1}{2}} + \lambda \bm{I})^{-1} \bm{R}^{\frac{1}{2}} \bm{H} \bm{k}_{\bm{x}}$$ where $\bm{H}=\bm{I}-\frac{1}{n}   \vec{\bm{1}}\vec{\bm{1}}^\top$, $\vec{\bm{1}}$ is a $n \times 1$ vector of ones and $\bm{R}$ is a diagonal re-weighting matrix with $[\bm{R}]_{i,i} = \frac{u(\bm{x}_i)}{p(\bm{x}_i)}$. The overall algorithm remains unchanged except for line 2 in Algorithm \ref{ouralgo} where the empirical estimate of the conditional kernel mean embeddings is replaced by its re-weighted version. We denote this variant of KIIM-HT as Rw-KIIM-HT where ``Rw'' stands for ``re-weighted''.

\section{Experiments}\label{experiments}
We describe our experimental setup and results on a synthetic dataset, a two-dimensional synthetic dataset and a real-world benchmark dataset T\"{u}bingen Cause-Effect Pairs (TCEP). We also conduct an experiment to analyze the sensitivity of the kernel based methods to the regularization parameter and an experiment to demonstrate the robustness of our method to different hyperparameters.
\subsection{Synthetic Data}
We test the effectiveness of our algorithm against a set of baseline on five sets of synthetic mappings from cause $x$ to effect $y$. These mappings include 1) ANM-1: $y = x^3 + x + \epsilon$, 2) ANM-2: $y = x + \epsilon$, 3) MNM-1: $y = (x^3 + x)\exp(\epsilon)$, 4) MNM-2: $(\sin(10 x) + \exp(3 x))\exp(\epsilon)$ and 5) CNM: $\log(x^6 + 5) + x^5 - \sin(x^2 \left| \epsilon \right|)$. We conduct experiments for two noise generating processes, where in the first process $\epsilon \sim \mathcal{N}(0, 1)$ is sampled from the standard normal distribution and in the second, $\epsilon \sim \mathcal{U}(0, 1)$ is sampled from the standard uniform distribution. The cause $x \sim \mathcal{N}(0, 1)$ is sampled from the standard normal distribution and a sample size of $100$ is used for each experimental instance. We compare our algorithm against ANM \cite{NIPS2008_3548}, IGCI \cite{Daniusis:2010:IDC:3023549.3023566}, KCDC \cite{Mitrovic:2018:CIV:3327757.3327802}, LiNGAM \cite{Shimizu:2006:LNA:1248547.1248619} and KIIM \cite{TODO_Zhitang_arxiv}. In KCDC, KIIM and KIIM-HT, both gram matrices $\bm{K}_x$ and $\bm{K}_y$ are taken to be the multiplication of the rational quadratic (RQ) kernel defined as $k(\bm{x}, \bm{y}) = 1-\frac{\|\bm{x} - \bm{y}\|_2^2}{\|\bm{x} - \bm{y}\|_2^2+1}$ and the RBF kernel as defined in the preliminaries section. The scale-length hyperparameter of the RBF kernel is selected using the median heuristic and the kernel regularization parameter is set to $\lambda =  10^{-3}$. In addition, KIIM-HT utilizes a projection matrix $\bm{W}_{\theta}(\cdot)$ parameterized by an input layer of size \texttt{input\_dim} by \texttt{n\_hidden} and then an output layer of size \texttt{n\_hidden} by \texttt{rank} $\times$ \texttt{n\_samples} (i.e., $n$) which then gets reshaped to form the projection matrix. The number of hidden units is set to $\text{\texttt{n\_hidden}} = 20$ and the rank of the projection matrix is set to $\text{\texttt{rank}} = 100$. A value of $\lambda_{\text{reg}} =  10^{-3}$ is used for the regularization hyperparameter. A rectified linear unit activation is used for the hidden layer while a linear activation is used for the output layer. We use the Adam optimizer \cite{Kingma2014AdamAM} with default parameters, a learning rate of $ 10^{-3}$ and $\mathcal{I} = 100$ gradient steps for optimizing the loss function. IGCI is tested for both settings where the reference measure parameter is set to the uniform reference measure and the Gaussian reference measure which are denoted IGCI-$\mathcal{N}$ and IGCI-$\mathcal{U}$, respectively. We ran five trials for each experimental setting and we report the average performance and the accompanying standard deviation (SD). \begin{table*}[t]
\centering
\caption{Performance on the synthetic dataset.}\smallskip
\label{experiment1}
\begin{tabular}{@{} l c c c c c c c c @{}}
\toprule
\multicolumn{2}{c}{} & \multicolumn{7}{c}{Models $(\%)$} \\
\cmidrule(l){3-9}
Function      &    $\epsilon \sim$      & ANM & IGCI-$\mathcal{N}$ & IGCI-$\mathcal{U}$ & KCDC & LiNGAM & KIIM & KIIM-HT \\ \midrule
ANM-1 & $\mathcal{N}$ & $ 99.6 \pm 0.49 $ & $ 97.8 \pm 0.98 $           &  $ 100.0 \pm 0.0 $             & $ 74.8 \pm 3.6 $  & $ 100.0 \pm 0.0 $   & $ 100.0 \pm 0.0 $  & $ 100.0 \pm 0.0 $ \\ %\midrule
      & $\mathcal{U}$  & $ 100.0 \pm 0.0 $ & $ 99.6 \pm 0.49 $            & $ 100.0 \pm 0.0 $              & $ 0.0 \pm 0.0 $    & $ 100.0 \pm 0.0 $   & $ 96.2 \pm 1.47 $   & $ 100.0 \pm 0.0 $ \\ \midrule
ANM-2 & $\mathcal{N}$ & $ 47.8 \pm 2.86 $ & $ 52.6 \pm 4.8 $         & $ 48.6 \pm 7.63 $              & $ 57.2 \pm 3.12 $    & $ 0.0 \pm 0.0 $    & $ 80.4 \pm 3.61 $   & $ 96.4 \pm 0.49 $ \\ %\midrule
      & $\mathcal{U}$  & $ 70.4 \pm 4.92 $ & $ 52.2 \pm 4.26 $          &  $ 48.4 \pm 5.75 $            & $ 48.4 \pm 7.81 $  & $ 0.0 \pm 0.0 $    & $ 48.8 \pm 4.26 $  & $ 67.2 \pm 3.12 $ \\ \midrule
MNM-1 & $\mathcal{N}$ & $ 0.0 \pm 0.0 $  & $ 100.0 \pm 0.0 $          &  $ 100.0 \pm 0.0 $              & $ 0.0 \pm 0.0 $   & $ 100.0 \pm 0.0 $   & $ 100.0 \pm 0.0 $  & $ 100.0 \pm 0.0 $ \\ %\midrule
      & $\mathcal{U}$  & $ 0.0 \pm 0.0 $  &  $ 99.8 \pm 0.4 $          & $ 100.0 \pm 0.0 $              & $ 0.0 \pm 0.0 $     & $ 100.0 \pm 0.0 $    & $ 99.6 \pm 0.49 $   & $ 100.0 \pm 0.0 $ \\ \midrule
MNM-2 & $\mathcal{N}$ & $ 2.4 \pm 1.02 $   & $ 100.0 \pm 0.0 $           & $ 100.0 \pm 0.0 $             & $ 0.0 \pm 0.0 $   & $ 100.0 \pm 0.0 $   & $ 100.0 \pm 0.0 $  & $ 100.0 \pm 0.0 $ \\ %\midrule
      & $\mathcal{U}$  & $ 29.2 \pm 6.65 $  &  $ 100.0 \pm 0.0 $           &  $ 100.0 \pm 0.0 $            & $ 0.0 \pm 0.0 $    & $ 100.0 \pm 0.0 $    & $ 99.2 \pm 1.17 $   & $ 100.0 \pm 0.0 $ \\ \midrule
CNM   & $\mathcal{N}$ &  $ 47.4 \pm 4.13 $  &  $ 100.0 \pm 0.0 $         & $ 100.0 \pm 0.0 $           & $ 0.0 \pm 0.0 $   & $ 100.0 \pm 0.0 $ & $ 100.0 \pm 0.0 $ & $ 100.0 \pm 0.0 $ \\ %\midrule
      & $\mathcal{U}$  & $ 30.8 \pm 5.56 $  & $ 100.0 \pm 0.0 $            &  $ 100.0 \pm 0.0 $             & $ 0.0 \pm 0.0 $  & $ 100.0 \pm 0.0 $    & $ 100.0 \pm 0.0 $  & $ 100.0 \pm 0.0 $ \\ \bottomrule
\end{tabular}
\end{table*}

The results are summarized in Table \ref{experiment1}. As expected, ANM performs well on the ANM-1 and ANM-2 datasets due to the models assumption of additive noise. However, this assumption also hampers its performance on data generating mechanisms involving nonadditive noise as evident in the model's poorer performance on the other datasets. IGCI shows robust performance with both uniform and Gaussian reference measure while on the other hand, we were unfortunately unable to reproduce the positive results of KCDC as reported in their paper. We conjecture KCDC may be sensitive to the regularization hyperparameter $\lambda$ which is unspecified in the original paper and we investigate this further in our later experiments. LiNGAM exhibits a performance in the extremes where it achieves perfect on most of the functional and noise settings but then there are also a few settings where it classifies with an accuracy of zero percent. KIIM also performs decently across all settings; compared to IGCI, one of the best baselines, KIIM significantly outperforms on $\epsilon \sim \mathcal{N}(0, 1)$ for ANM-2 while it arguably matches the performance for all other settings (i.e., slight advantage for IGCI on $\epsilon \sim \mathcal{U}(0, 1)$ for MNM-1 and MNM-2). Similarly, KIIM-HT displays very strong performance; for all settings, KIIM-HT either matches or outperforms all other baselines except for the setting $\epsilon \sim \mathcal{U}(0, 1)$ for ANM-2 where KIIM-HT ranks second behind ANM. KIIM-HT also shows robustness in the stability of the algorithm as it has the least variance on the ANM-2 settings which were the hardest and had the most varied performances.
\subsection{Two-Dimensional Synthetic Data}
We examine the performance of our algorithm on a set of two-dimensional vector valued cause-effect pairs. Indeed, real-world statistical experiments oftentimes measure multiple predictor and response variables and it would be unusual to observe the value of only one random variable. There are ten sets of synthetic mappings between cause $\bm{x}$ and effect $\bm{y}$ which are generated by taking pairwise combinations of the scalar mappings of the previous experiment and then collating the respective causes and effects to form a vector (i.e., $\bm{x}_i$ and the associated $\bm{y}_i$ corresponds to an independent scalar mapping where $i \in \{1, 2\}$). We abstain from permuting the noise process of each dimension and instead each vector mapping is associated with noise generated from either the standard normal or standard uniform distribution, resulting in a total of twenty settings. We use a sample size of $n = 5$ while all other experimental setup parameters remain the same. Although theoretically, there are no prohibitions on extending ANM and LiNGAM to operate on vector-valued inputs, we were unable to find such implementations and hence, we only compare against IGCI, KCDC and KIIM. \begin{table*}[t]
\centering
\caption{Performance on the two-dimensional synthetic dataset.}\smallskip
\label{experiment2}
\begin{tabular}{@{} l c c c c c c @{}}
\toprule
\multicolumn{2}{c}{} & \multicolumn{5}{c}{Models $(\%)$} \\
\cmidrule(l){3-7}
%\toprule
Functions            &    $\epsilon \sim$      & IGCI-$\mathcal{N}$ & IGCI-$\mathcal{U}$ & KCDC & KIIM & KIIM-HT \\ \midrule
ANM-1 ANM-2 & $\mathcal{N}$ & $ 65.8 \pm 5.04 $ & $ 72.4 \pm 4.63 $ & $ 18.0 \pm 4.05 $ & $ 8.0 \pm 0.89 $ & $ 92.0 \pm 2.28 $ \\ %\midrule
            & $\mathcal{U}$  & $ 70.2 \pm 2.4 $ & $ 77.0 \pm 4.34 $ & $ 21.6 \pm 3.38 $ & $ 10.4 \pm 2.87 $ & $ 94.0 \pm 1.67 $ \\ \midrule
ANM-1 MNM-1 & $\mathcal{N}$ & $ 75.6 \pm 4.54 $ & $ 82.6 \pm 2.94 $ & $ 9.2 \pm 1.17 $ & $ 2.8 \pm 1.17 $ & $ 98.0 \pm 1.1 $ \\ %\midrule
            & $\mathcal{U}$  & $ 76.4 \pm 4.13 $ & $ 86.0 \pm 2.1 $ & $ 3.2 \pm 2.48 $ & $ 0.4 \pm 0.49 $ & $ 100.0 \pm 0.0 $ \\ \midrule
ANM-1 MNM-2 & $\mathcal{N}$ & $ 88.4 \pm 3.26 $ & $ 93.6 \pm 1.62 $ & $ 5.4 \pm 1.02 $ & $ 1.8 \pm 0.4 $ & $ 98.2 \pm 0.4 $ \\ %\midrule
            & $\mathcal{U}$  & $ 89.4 \pm 2.8 $ & $ 92.8 \pm 2.23 $ & $ 5.6 \pm 2.42 $ & $ 1.8 \pm 1.72 $ & $ 99.0 \pm 0.63 $ \\ \midrule
ANM-1 CNM   & $\mathcal{N}$ & $ 87.8 \pm 2.4 $ & $ 94.0 \pm 1.79 $ & $ 25.2 \pm 2.71 $ & $ 9.8 \pm 2.71 $ & $ 99.2 \pm 0.98 $ \\ %\midrule
            & $\mathcal{U}$  & $ 91.2 \pm 1.47 $ & $ 95.2 \pm 0.98 $ & $ 29.0 \pm 2.76 $ & $ 11.2 \pm 4.87 $ & $ 100.0 \pm 0.0 $ \\ \midrule
ANM-2 MNM-1 & $\mathcal{N}$ & $ 72.6 \pm 2.24 $ & $ 78.4 \pm 3.61 $ & $ 20.2 \pm 4.17 $ & $ 11.0 \pm 0.89 $ & $ 89.2 \pm 3.71 $ \\ %\midrule
            & $\mathcal{U}$  & $ 77.8 \pm 3.19 $ & $ 84.2 \pm 3.87 $ & $ 9.8 \pm 2.14 $ & $ 4.8 \pm 1.47 $ & $ 94.6 \pm 1.85 $ \\ \midrule
ANM-2 MNM-2 & $\mathcal{N}$ & $ 88.6 \pm 1.85 $ & $ 92.0 \pm 1.26 $ & $ 18.0 \pm 1.79 $ & $ 7.6 \pm 1.2 $ & $ 95.6 \pm 2.06 $ \\ %\midrule
            & $\mathcal{U}$  & $ 90.0 \pm 3.52 $ & $ 91.2 \pm 1.47 $ & $ 9.0 \pm 2.61 $ & $ 5.4 \pm 2.8 $ & $ 96.6 \pm 1.85 $ \\ \midrule
ANM-2 CNM   & $\mathcal{N}$ & $ 86.0 \pm 4.05 $ & $ 90.0 \pm 1.79 $ & $ 42.0 \pm 6.48 $ & $ 24.0 \pm 2.53 $ & $ 92.8 \pm 1.17 $ \\ %\midrule
            & $\mathcal{U}$  & $ 91.0 \pm 3.16 $ & $ 91.2 \pm 1.47 $ & $ 56.4 \pm 1.85 $ & $ 38.2 \pm 4.96 $ & $ 94.6 \pm 2.24 $ \\ \midrule
MNM-1 MNM-2 & $\mathcal{N}$ & $ 93.2 \pm 1.17 $ & $ 94.0 \pm 2.61 $ & $ 11.2 \pm 2.14 $ & $ 3.6 \pm 1.02 $ & $ 98.6 \pm 0.49 $ \\ %\midrule
            & $\mathcal{U}$  & $ 91.8 \pm 3.12 $ & $ 91.6 \pm 1.85 $ & $ 2.6 \pm 0.8 $ & $ 0.0 \pm 0.0 $ & $ 100.0 \pm 0.0 $ \\ \midrule
MNM-1 CNM   & $\mathcal{N}$ & $ 93.4 \pm 3.38 $ & $ 94.4 \pm 1.36 $ & $ 29.4 \pm 4.13 $ & $ 14.8 \pm 1.6 $ & $ 99.2 \pm 0.75 $ \\ %\midrule
            & $\mathcal{U}$  & $ 90.8 \pm 1.17 $ & $ 95.0 \pm 1.79 $ & $ 18.8 \pm 3.71 $ & $ 3.8 \pm 1.47 $ & $ 99.8 \pm 0.4 $ \\ \midrule
MNM-2 CNM   & $\mathcal{N}$ & $ 96.0 \pm 1.41 $ & $ 96.4 \pm 2.24 $ & $ 21.8 \pm 5.78 $ & $ 9.0 \pm 1.79 $ & $ 99.4 \pm 0.8 $ \\ %\midrule
            & $\mathcal{U}$  & $ 97.2 \pm 1.33 $ & $ 97.2 \pm 1.17 $ & $ 18.0 \pm 3.03 $ & $ 3.6 \pm 0.8 $ & $ 99.4 \pm 0.8 $ \\ \bottomrule
\end{tabular}
\end{table*} The results are summarized in Table \ref{experiment2}. We see again that both IGCI-$\mathcal{N}$ and IGCI-$\mathcal{U}$ are quite robust, consistently achieving performances in the $80$s and $90$s with only a few in the $70$s. Similarly, KCDC also remains consistent with previous findings, showing subpar performance with the best performing setting at $56.4$ percent. KIIM, on the other hand, deviates from previous results and performs very poorly on the two-dimensional settings, with the best performing setting only at $38.2$ percent. We are unsure of KIIM's large difference in performance across the two experiments, but it appears that KIIM may not perform well under a low-data regime and experimental evidence, included in Appendix A, corroborates this as the performance of KIIM improves with increased data samples. We conjecture KIIM's sensitivity to sample size is due to its usage of eigendecomposition as a core part in its algorithm. Decomposing a data matrix into a set of eigvenvalues and eigenvectors could be viewed as a data summarization technique expressed through said eigenpairs and indeed, one of the most elementary dimensionality reduction technique, PCA, amounts to this procedure. Of course, a summarization is logical when there are a lot of data samples, but a summary is an antithesis in a low-data regime, which may explain why KIIM is doing so poorly when $n = 5$ as in such a setting, it may by operating on nonsense returned by eigendecomposition. Lastly, we see that KIIM-HT does the best, beating all other baselines, including IGCI, in every single instance. The accuracy of KIIM-HT is consistently in the high $90$s and it achieves no lower than $91$ percent in all the settings. Similar to the previous experiment, KIIM-HT shows significantly low variability in all of the settings; the only clear loss for the model is on ANM-2 MNM-2 for $\epsilon \sim \mathcal{N}(0, 1)$ while for all other settings KIIM-HT consistently has the lowest variability.
\subsection{T\"{u}bingen Cause-Effect Pairs}
In this experiment we test our algorithm on real-world data. We use the open dataset T\"{u}bingen Cause-Effect Pairs (TCEP) \cite{Mooij:2016:DCE:2946645.2946677} which is a popular benchmark consisting of approximately $108$ (varying between versions) cause and effect pairs collected in various real-world domains (e.g., meteorology, biology, engineering, etc.) with the ground-truth causal direction labelled by human experts. We use the version of TCEP provided in the Causal Discovery Toolbox open-source Python package \cite{cdt} which delivers the data already preprocessed e.g., removal of multivariate pairs (not necessary for all models as evident in previous section but rather for fair comparisons to previous work) and arrangement of variables so that $x$ and $y$ is always cause and effect, respectively. Additionally, some pairs exhibit significantly large sample size (i.e., the largest sample size of all the pairs is $16382$) and it is well-known that kernel methods, at least when naively implemented, does not scale well with large training sets. Therefore, to cater to our kernel-based method, in each experimental instance, the training set is reduced via randomly sampling the first $400$ samples from the original set. For fair comparisons we subject all models to this downsampling procedure and we show in Appendix B that this did not adversely affect the non-kernel based methods. As aforementioned, real-world data are rife with noise and outliers and therefore, we use the re-weighting schemes for KIIM and KIIM-HT to mitigate their effects. We test two different reference distributions where one is the Gaussian distribution and the other is the Laplace distribution which we denote, respectively, by appending a ``-$\mathcal{N}$'' and ``-$\mathcal{L}$'' to the model name. \begin{table}[t]
\centering
\caption{Performance on TCEP.}\smallskip
\label{experiment3}
\begin{tabular}{@{} l r @{}}
\toprule
Method    & TCEP $(\%)$ \\ \midrule
ANM       & $53.4 \pm 2.42$ \\ %\midrule
IGCI-$\mathcal{N}$    & $61.6 \pm 0.49$ \\ %\midrule
IGCI-$\mathcal{U}$    & $65.0 \pm 1.26$ \\ %\midrule
KCDC      & $59.8 \pm 1.72$ \\ %\midrule
LiNGAM    & $34.2 \pm 1.60$ \\ %\midrule
KIIM    & $40.6 \pm 2.33$ \\ %\midrule
Rw-KIIM-$\mathcal{N}$    & $69.8 \pm 1.94$ \\ %\midrule
Rw-KIIM-$\mathcal{L}$    & $62.4 \pm 0.80$ \\ %\midrule
KIIM-HT    & $37.8 \pm 0.75$ \\ %\midrule
Rw-KIIM-HT-$\mathcal{N}$ & $71.2 \pm 1.72$ \\ %\midrule %72,73,68,71,72
Rw-KIIM-HT-$\mathcal{L}$ & $69.0 \pm 2.1$ \\ \bottomrule %67,67,72,68,71
\end{tabular}
\end{table} Table \ref{experiment3} provides a summary of the results. ANM, IGCI-$\mathcal{N}$ and IGCI-$\mathcal{U}$ have similar performances, consistent with previous results. LiNGAM does poorly while KCDC performs just behind IGCI, consistent with results reported in one of the previous works. As expected, both vanilla KIIM and KIIM-HT perform poorly due to the presence of noise and outliers in real-world data, whereas re-weighted KIIM and its heterogeneous extension, in general, does well, with KIIM-HT-$\mathcal{L}$ performing the best.
%is excellent, consistent with results reported in the original paper. Reweighted KIIM and its nonlinear extension, in general, performs well, with KIIM-HT-$\mathcal{L}$ performing the best.
\subsection{Sensitivity to $\lambda$}
We examine the impact of the kernel regularization parameter $\lambda$ inherent to the kernel based methods (e.g., KCDC, KIIM and KIIM-HT). The motivation for this experiment partially stems from our inability to replicate the results of KCDC and we suspect that KCDC's performance may vary significantly with $\lambda$ which is a value that is unspecified in the experiments of the original paper. We test the performance of the kernel based methods for $\lambda \in \{ 10^{-3},  10^{-2},  10^{-1}, 1, 5, 10, 50\}$ on the synthetic and the two-dimensional synthetic dataset. We additionally test KCDC on the two-dimensional synthetic dataset with $100$ samples. The results are given in Appendix C. For the the synthetic dataset, the performance of KCDC is positively correlated, peaking and maintaining thereafter at $\lambda = 1$ while KIIM and KIIM-HT both exhibit performances invariant to $\lambda$. For the two-dimensional dataset, KCDC is similarly positively correlated with $\lambda$ whereas KIIM is also positively correlated while KIIM-HT is negatively correlated. With $100$ samples, KCDC is also positively correlated with $\lambda$, doing poorly in the beginning, performing perfect at $\lambda = 10^{-1}$ then declining thereafter. The evidence is in favor of our hypothesis that KCDC's strong performance may be due to large $\lambda$. We stress that such a setting would be unusual because the main purpose of $\lambda$ is to regularize the Gram matrix and an unduly large $\lambda$ ought to interfere with the data. Furthermore, the effects of a large $\lambda$ is magnified for KCDC relative to KIIM and KIIM-HT as the former uses a regularization term given by $n \lambda$ as opposed to just $\lambda$ for the latter two.

%We examine the impact of the kernel regularization parameter $\lambda$ inherent to the kernel based methods (e.g., KCDC, KIIM and KIIM-HT). The motivation for this experiment partially stems from our inability to replicate the results of KCDC and we suspect that KCDC's performance may vary significantly with $\lambda$ which is a value that is unspecified in the experiments of the original paper. We test the performance of the kernel based methods for $\lambda \in \{ 10^{-3},  10^{-2},  10^{-1}, 1, 5, 10, 50\}$ on the synthetic and the two-dimensional synthetic dataset. The results are given in Appendix C. For the the synthetic dataset, the performance of KCDC is positively correlated, peaking and maintaining thereafter at $\lambda = 1$ while KIIM and KIIM-HT both exhibit performances invariant to $\lambda$. For the two-dimensional dataset, KCDC is similarly positively correlated with $\lambda$ whereas KIIM is also positively correlated while KIIM-HT is negatively correlated. The evidence is in favor of our hypothesis that KCDC's strong performance may be due to large $\lambda$. We stress that such a setting would be unusual because the main purpose of $\lambda$ is to regularized the Gram matrix and an unduly large $\lambda$ ought to interfere with the data. Furthermore, the effects of a large $\lambda$ is magnified for KCDC relative to KIIM and KIIM-HT as the former uses a regularization term given by $n \lambda$ as opposed to just $\lambda$ for the latter two.

\subsection{Robustness to Hyperparameters}
Compared to KCDC and KIIM, KIIM-HT introduces a few additional tunable hyperparameters and in this last experiment, we demonstrate the robustness of our method to different hyperparameter values. We focus on the rank of the projection matrix $r$ and the regularization hyperparameter $\lambda_{\text{reg}}$. A straightforward approach is taken where we perform a grid search of the hyperparameters over the values $\mathbb{A}_r \times \mathbb{B}_{\lambda_{\text{reg}}}$ where $\mathbb{A}_r = \{5, 10, 20, 80, 100\}$ and $\mathbb{B}_{\lambda_{\text{reg}}} = \{ 10^{-4},  10^{-3},  10^{-2},  10^{-1}, 1\}$ is the set of values we search over $r$ and $\lambda_{\text{ref}}$, respectively. For each hyperparameter combination, we run the suite of experiments presented in the synthetic data section as well as the two-dimensional synthetic data section. The results are given in Appendix D. For both experimental suites, we see that the results are very consistent. In the first suite of experiments, we obtain unvarying results of one hundred across all experimental settings except for ANM-2 where with $\epsilon \sim \mathcal{N}(0, 1)$ we obtain a mean performance of $94.04$ with SD $4.30$ while for $\epsilon \sim \mathcal{U}(0, 1)$ a mean of $64.92$ with SD $5.18$, displaying quite robust performances. The second suite of experiments similarly displays consistent performances with most of the figures in the 90s and just a few in the 80s and 70s.

\section{Conclusion}\label{conclusion}
In this work, we focus on the problem of discovering the direction of causation of two potentially multivariate random variables. We discuss some limitations of current state-of-the-art methods where specifically we show that a global linear projection of the conditional mean embedding is an insufficient metric to compare the intrinsic invariance/deviance of the conditional distributions. To rectify this, we propose Kernel Intrinsic Invariance Measure with Heterogeneous Transformation (KIIM-HT) which uses an artificial neural network trained via stochastic gradient descent to heterogeneously and locally project each conditional mean embedding into a subspace where only relevant higher-order statistics are extracted for causal discovery. Results and comparisons on a synthetic dataset, a two-dimensional synthetic dataset and a real-world dataset verify the effectiveness of our approach while a sensitivity analysis to the regularization parameter attempts at a candid comparison to previous work and finally, an experiment with trials on various hyperparameter values demonstrate the robustness of our algorithm.

\bibliography{example_paper}
\bibliographystyle{icml2020}

\end{document}